\documentclass[11pt,a4paper]{article}
\usepackage[hyperref]{emnlp2020}
\usepackage{times}
\usepackage{latexsym}
\usepackage{graphicx}
\usepackage{diagbox}
\usepackage{enumitem}
\usepackage{multirow}
\usepackage{indentfirst}
\usepackage{colortbl}

\usepackage{microtype}

\aclfinalcopy 
\def\aclpaperid{1990} 

\title{Zero-shot Entity Linking with Efficient Long Range Sequence Modeling}

\author{Zonghai Yao \\ University of Massachusetts, Amherst \\  \texttt{zonghaiyao@cs.umass.edu}
        \AND
        Liangliang Cao \\ University of Massachusetts, Amherst \\ \texttt{llcao@cs.umass.edu} \And
        Huapu Pan \\ Google Research \\ \texttt{huapupan@google.com}}

\date{}

\begin{document}
\maketitle
\begin{abstract}
This paper considers the problem of zero-shot entity linking, in which a link in the test time may not present in training.
Following the prevailing BERT-based research efforts, we find a simple yet effective way is to expand the long-range sequence modeling. 
Unlike many previous methods, our method does not require expensive pre-training of BERT with long position embeddings. 
Instead, we propose an efficient position embeddings initialization method called Embedding-repeat, which initializes larger position embeddings based on BERT-Base. On Wikia's zero-shot EL dataset, our method improves the SOTA from 76.06\% to 79.08\%, and for its long data, the corresponding improvement is from 74.57\% to 82.14\%.
Our experiments suggest the effectiveness of long-range sequence modeling without retraining the BERT model. \footnote{Our code are publicly available. \url{https://github.com/seasonyao/Zero-Shot-Entity-Linking}.}

\end{abstract}

\section{Introduction}
\begin{figure}[t]
        \centering
        \includegraphics[width=\linewidth]{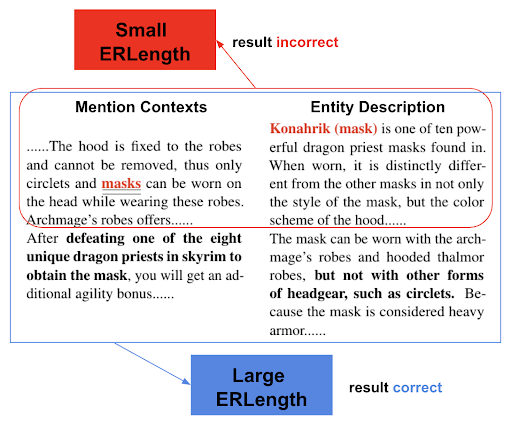}
        \caption{Only models with large ERLength can solve this entity linking problem because only they can get valuable critical information in the mention contexts and entity description.}
        \label{example}
    \end{figure}

    Entity linking (EL) is the task of grounding entity mentions by linking them to entries in a given database or dictionary of entities. Traditional EL approaches often assume that entities linked at the test time are present in the training set. Nevertheless, many real-world applications prefer the zero-shot setting, where there is no external knowledge and a short text description provides the only information we have for each entity \cite{sil2012linking,wang2015language}. For zero-shot entity linking \cite{logeswaran2019zero}, it is crucial to consider the context of entity description and mention, so that the system can generalize to unseen entities. However, most of the BERT-based models are based on a context window with 512 tokens, limited to capturing the long-range of context. This paper defines a model's \textbf{Effective-Reading-Length (ERLength)} as the total length of the mention contexts and entity description that it can read. Figure \ref{example} demonstrates an example where long ERlengths are more preferred than short ones.
    
    \indent Many existing methods that can be used to expand ERLength \cite{sohoni2019low, dai2019transformer}, however, often need to completely re-do pre-training with the masked language modeling objective on the vast general corpus (like Wikipedia), which is not only very expensive but also impossible in many scenarios.\\
    \indent This paper proposes a practical way, Embeddings-repeat, to expand BERT's ERLength by initializing larger position embeddings, allowing reading all information in the context. Note our method differs from previous works since it can directly use the larger position embeddings initialized from BERT-Base to do fine-tuning on downstream tasks without any retraining. Extensive experiments are conducted to compare different ways of expanding ERLength, and the results show that Embeddings-repeat can robustly improve performance. Most importantly, we improve the accuracy from 76.06\% to 79.08\% in Wikia's zero-shot EL dataset, from 74.57\% to 82.14\% for its long data. Since our method is effective and easy to implement, we expect our method will be useful for other downstream NLP tasks.

\section{Related work}
    \paragraph{Zero-shot Entity Linking} Most state-of-the-art entity linking methods are composed of two steps: candidate generation  \cite{sil2012linking, vilnis2018hierarchical, radford2018improving} and candidate ranking \cite{he2013learning, sun2015modeling, yamada2016joint}. 
    \citet{logeswaran2019zero}  proposed the zero-shot entity linking task, where mentions must be linked to unseen entities without in-domain labeled data.
    For each mention, the model first uses BM25 \cite{robertson2009probabilistic} to generate 64 candidates. For each candidate, BERT \cite{devlin2018bert} will read a sequence pair combining mention contexts and entity description and produce a vector representation for it. Then, the model will rank the candidates based on these vectors.
    This paper discusses how to improve \citet{logeswaran2019zero} by efficiently expanding the ERLength.
    
    \paragraph{Modeling long documents} 
    The simplest way to work around the 512 limit is to truncate the document\cite{xie2019unsupervised, liu2019roberta}. It suffers from severe information loss, which does not meet sufficient information in the zero-shot entity linking. 
    Recently there has been an explosive amount of efforts to improve long-range sequence modeling \cite{sukhbaatar2019adaptive, rae2019compressive, child2019generating, ye2019bp, qiu2019blockwise, lample2019large, lan2019albert}. However, they all need to initialize new position embeddings and do expensive retraining on the general corpus (like Wikipedia) to learn the positional relationship in longer documents before fine-tuning downstream tasks. Moreover, the exploration of the impact of long-range sequence modeling on entity linking is still blank. So in this study, we will explore a different approach, which initializes larger position embeddings based on the existing small one in BERT-Base, and can be used directly in the fine-tuning without expensive retraining.
        
        \begin{table*}
            \centering
            \begin{tabular}{l|c|cccccc|c}
            \hline
            \multirow{2}*{Method} & \multirow{2}*{ERLength} & \multicolumn{6}{|c|}{eval} & \multirow{2}*{test} \\
            &  & $set_1$ & $set_2$ & $set_3$ & $set_4$ & avg  & long &  \\
            \hline
            \citet{logeswaran2019zero} & 256 & \small{83.40} & \small{79.00} & \small{73.03} & \small{68.82} & 76.06 & 74.57 & 75.06\\
            
            BERT  & 512 & \small{83.45} & \small{80.03} & \small{71.88} & \small{72.53} & 76.97 & 78.54 & -\\
            
            \citet{logeswaran2019zero} (DAP) & 256 &  \small{82.82} & \small{81.59} & \small{75.34} & \small{72.52} & 78.07 & 76.89 & 77.05\\
            
            $E_{repeat}$ & 1024 & \small{87.02} & \small{81.52} & \small{73.48} & \small{74.37} & 79.08 & 82.14 & 77.58\\
            
            $E_{repeat}$ + DAP & 1024 & \small\textbf{89.67} & \small\textbf{83.53} & \small\textbf{75.37} & \small\textbf{74.96} & \textbf{80.88} & \textbf{82.14} & \textbf{79.64}\\
            \hline
            \end{tabular}
            \caption{Our methods with long ERlength outperform state of the art. Especially, the accuracy of the long data increases from 74.57\% to 82.14\% compared with the benchmark. Here, we call all data whose DLength exceeds 512 (the maximum number  BERT  can read) as \textbf{long} data. If we also use DAP, the best accuracy is 80.88\% in validation data, and 79.64\% in test data. Note: $set_1$: Coronation street, $set_2$: Muppets, $set_3$: Ice hockey, $set_4$: Elder scrolls, DAP: Domain Adaptive Pre-training \cite{logeswaran2019zero}.}
            \label{baseline}
        \end{table*}

\section{Method}
    \subsection{Overview}
     \begin{figure}[ht]
            \centering
            \includegraphics[width=\linewidth, trim = {0 0 0 0}, clip]{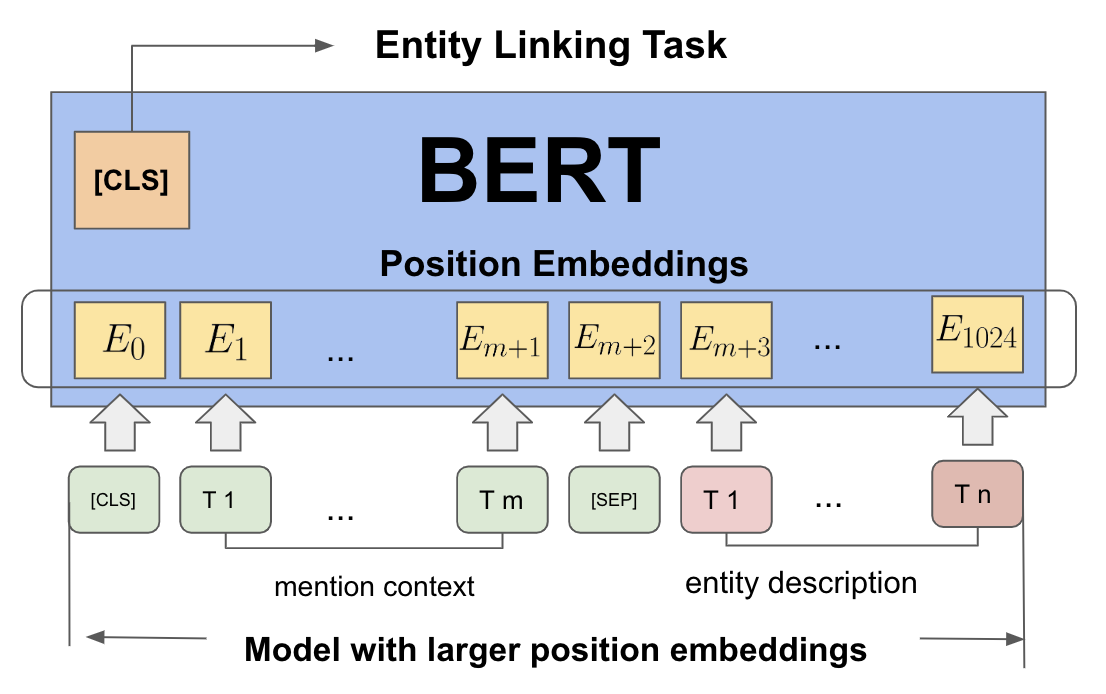}
            \caption{BERT doing entity linking with larger position embeddings}
            \label{bert_model}
    \end{figure}
    Figure \ref{bert_model} describes how to use BERT for zero-shot entity linking tasks with larger position embeddings. Following \citet{logeswaran2019zero}, we adopt a two-stage pipeline consisting of a fast candidate generation stage, followed by a more expensive but powerful candidate ranking stage \cite{ganea2017deep, kolitsas2018end, wu2019zero}. We use BM25 for the candidate generation stage and get 64 candidate entities for every mention. For the candidate ranking stage, as in BERT, the mention contexts $m$ and candidate entity description $e$ are concatenated as a sequence pair together with special start and separator tokens: ([CLS] $m$ [SEP] $e$ [SEP]). The Transformer \cite{vaswani2017attention} will encode this sequence pair, and the position embeddings inside will capture the position information of individual words. At the last hidden layer, the Transformer produces a vector representation $h_{m,e}$ of the input pair through the special pooling token [CLS]. And then entities in a given candidate set are scored as $softmax(\omega^\top{h}_{m,e})$ where $\omega$ is a learned parameter vector. \\
    \indent Since the size of position embeddings is limited to 512 in BERT, how to capture position information beyond this size is what we hope to improve. In general, for new and larger position embeddings, we often need to re-initialize it with the larger size, and then retrain on general corpus like Wikipedia to learn the positional relationship in longer documents. However, we found that the relationship between different positions in the text is related. We can initialize larger position embeddings from the small ones in BERT-Base, and then without any expensive retraining, directly use it to complete the fine-tuning on the downstream tasks.

    \subsection{Position embeddings initialization}
    
        \begin{figure}[ht]
            \centering
            \includegraphics[width=\linewidth]{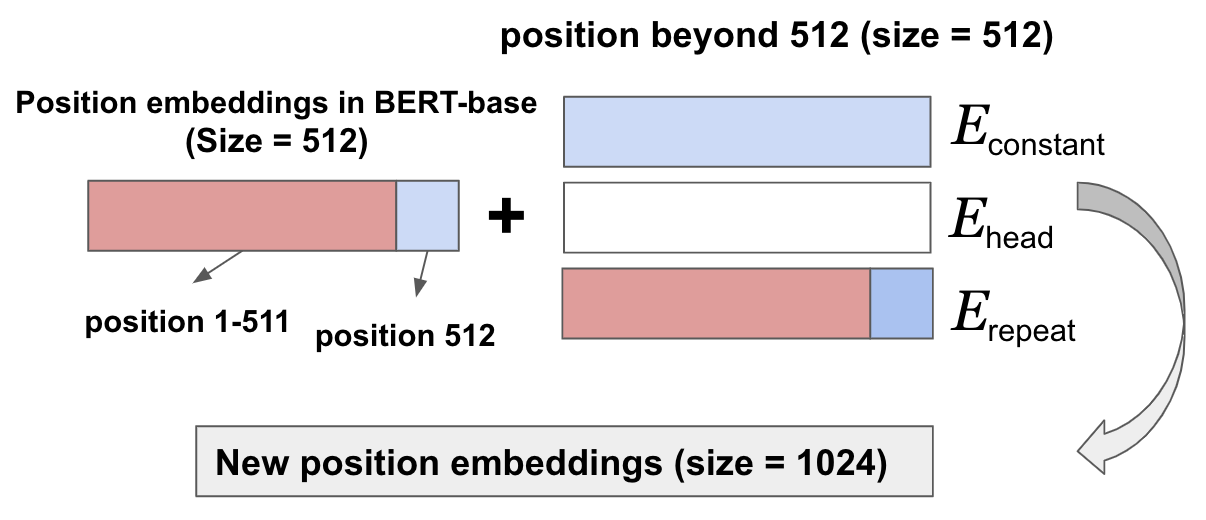}
            \caption{BERT model with larger position embeddings which initialized from different method}
            \label{model}
        \end{figure}
        
       It is reasonable to assume that the larger position embeddings have similar first 512 values with the small one since they all express the corresponding relationship between tokens when the input length is less than 512. For those positions over 512, we introduce a particular method \textbf{Embeddings-repeat ($E_{repeat}$)} to initialize larger position embeddings by repeating the small one from BERT-Base as analysis of BERT’s attention heads shows a strong learned bias to attend to the local context, including the previous or next token \cite{clark2019does}. We assume using $E_{repeat}$ preserves this local structure everywhere except at the partition boundaries. For example, for a 1024 position embeddings model, we will initialize the first 512 positions and the last 512 positions, respectively, from BERT-Base. \\
       \indent To verify the rationality of $E_{repeat}$, we also proposed two other methods as the comparison. $E_{head}$ assumes only the first 512 positions in the larger position embeddings are similar to that in the small one, so it initializes the first 512 positions from BERT-Base and randomly initializes those exceeding 512. $E_{constant}$ also uses position embeddings in BERT-Base to initialize its first 512 positions. However, it uses the value of position 512 to initialize those exceeding 512, since it assumes the relationship between two tokens over a long distance tend to be constant. In the following experimental part, we show that at least in this task, using $E_{repeat}$ to expand the ERLength of BERT is most effective.

\section{Experiments}

    \begin{table*}[!htbp]
          \centering
          \resizebox{\textwidth}{24mm}{
          \begin{tabular}{|c|c|c|c|c|c|c|c|c|c|c|c|}
          \hline
          \multicolumn{12}{|c|}{Proportion of data in different DLength intervals}\\
          \hline
          \cline{3-12}
          \multicolumn{2}{|c|}{\multirow{2}*{DLength}}&\multirow{2}*{(0,200)}&\multirow{2}*{[200,300)}&\multirow{2}*{[300,400)}&\multirow{2}*{[400,500)}&\multirow{2}*{[500,600)}&\multirow{2}*{[600,700)}&\multirow{2}*{[700,800)}&\multirow{2}*{[800,900)}&\multirow{2}*{[900,1000)}&\multirow{2}*{[1000,$+\infty$)}\\
          \multicolumn{2}{|c|}{}&&&&&&&&&&\\
          \hline
          \cline{3-12}
          \multicolumn{2}{|c|}{$\%$ Of total}& 10.62 & 14.62 & 11.92 & 9.96 & 15.18 & 14.11 & 8.56 & 4.91 & 3.39 & 6.83 \\
          \hline
          \multicolumn{12}{|c|}{Accuracy for model with different ERLength on different DLength interval}\\
          \hline
          \multirow{7}*{\rotatebox{90}{ERLength}}
          &64 & 60.76 & 62.10 & 62.20 & 58.61 & 64.18 & 61.50 & 62.45 & 60.52 & 65.41 & 62.57 \\
          \cline{2-12}
          &128 & 73.57 & 70.57 & 71.31 & 67.64 & 72.34 & 69.34 & 70.87 & 70.12 & 68.49 & 71.69 \\
          \cline{2-12}
          &256 & \textcolor{red}{75.52} & 74.16 & 75.50 & 73.34 & 75.35 & 75.47 & 75.11 & 73.71 & 77 & 74.34 \\
          \cline{2-12}
          &384 & \textcolor{red}{75.72} & \textcolor{red}{77.11} & \textcolor{red}{78.44} & \textcolor{red}{74.03} & \textcolor{red}{78.00} & 75.84 & 77.07 & 78.44 & 75.51 & 78.96 \\
          \cline{2-12}
          &512 & \textcolor{red}{76.64} & \textcolor{red}{75.81} & \textcolor{red}{78.00} & \textcolor{red}{74.68} & \textcolor{red}{77.99} & \textcolor{red}{77.33} & \textcolor{red}{78.97} & \textcolor{red}{81.34} & 75.48 & 79.49 \\
          \cline{2-12}
          &768 & \textcolor{red}{75.56} & \textcolor{red}{75.66} & \textcolor{red}{79.08} & \textcolor{red}{75.15} & \textcolor{red}{77.87} & \textcolor{red}{78.54} & \textcolor{red}{78.54} & \textcolor{red}{82.11} & 78.22 & 79.96 \\
          \cline{2-12}
          &1024 & \textcolor{red}{75.80} & \textcolor{red}{76.40} & \textcolor{red}{80.54} & \textcolor{red}{75.47} & \textcolor{red}{77.7} & \textcolor{red}{77.51} & \textcolor{red}{79.65} & \textcolor{red}{81.60} & \textcolor{red}{81.27} & \textcolor{red}{83.58} \\
          \hline
          \end{tabular}}
          \caption{The table shows the proportion of different DLength data and the accuracy of different ERLength models on different DLength data. Red represents the accuracy of the leading echelon in certain DLength data. It shows a cascading downward trend, which means that for larger DLength data, only models with larger ERLength can perform well, and even if ERLength is much larger than DLength, accuracy will not decline.}
          \label{impactOfERLength}
    \end{table*}
    
    \subsection{Dataset and experiment setup}
        We use Wikia's zero-shot EL dataset constructed by \citet{logeswaran2019zero}, which to our knowledge, is the best zero-shot EL benchmark. 
        To show the importance of long-range sequence modeling, we define the data's \textbf{DLength} as the total length of the mention contexts and entity description and examine the distribution of DLength on the dataset. As shown in Table \ref{impactOfERLength}, We found about half of the data have a DLength exceeding 512 tokens. Furthermore, $93\%$ of them are less than 1024. So we set the model's ERLength range from 0 to 1024, with which we explore how continuously expanding the model's ERLength will affect its performance on Wikia's zero-shot EL dataset. When we increase ERLength, we will assign the same size growth to the mention contexts and entity description, which we find is the most reasonable through our related experiments.\\
        
        \indent For all experiments, we follow the most recent work in studying zero-shot entity linking. We use the BERT-Base model architecture in all our experiments. The Masked LM objective \cite{devlin2018bert} is used for unsupervised pre-training. For fine-tuning language models (in the case of multi-stage pre-training) and fine-tuning on the Entity-Linking task, we use a small learning rate of 2e-5, following the recommendations from \citet{devlin2018bert}. All models are implemented in Tensorflow and optimized with Adam. All experiments were conducted with v3-8 TPU on Google Cloud.\\
        
        \indent Like \citet{logeswaran2019zero}, our entity linking performance is evaluated on the subset of test instances for which the gold entity is among the top-k candidates retrieved during candidate generation. Our IR-based candidate generation has a top-64 recall of 76\% and 68\% on the validation and test sets, respectively. Strengthening the candidate generation stage improves the final performance, but this is outside our work scope. Average performance across a set of domains is computed by macro-averaging. Performance is defined as the accuracy of the single-best identified entity (top-1 accuracy).\\

    \subsection{Comparison of different initializations}
    
        \begin{figure}[ht]
            \centering
            \includegraphics[width=\linewidth, trim = {0 0 0 0}, clip]{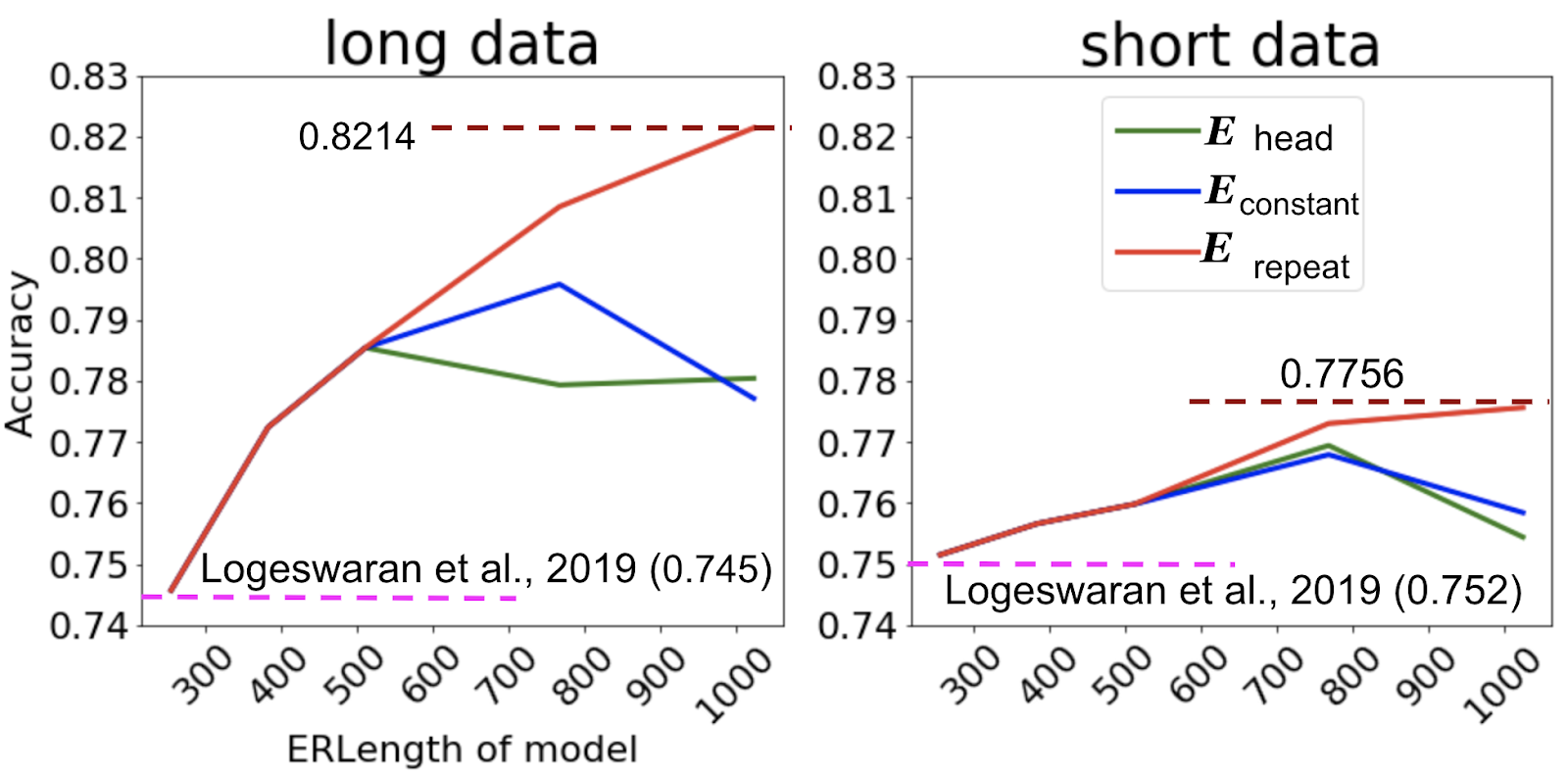}
            \caption{The accuracy of the model with different position embeddings initialization methods in long and short data. Note: We call all data whose DLength exceeds 512 as \textbf{long} data, otherwise, \textbf{short} data.}
            \label{initialization method}
        \end{figure}
        
        \indent The results of different position embeddings initialization methods are shown in figure \ref{initialization method}. It can be found that for both long and short data, $E_{repeat}$ has achieved the best results, especially its performance on long data is impressive. When the model's ERLength exceeds 512, only using $E_{head}$ produces worse results, which shows the importance of using the information of the first 512 positions to initialize the latter part. The model with $E_{constant}$ starts to decrease after its ERLength reaches about 768, which shows that its assumption is only reasonable when the model's ERLength is less than 768. Only when using $E_{repeat}$ to initialize we will see a stable and continuous improvement, which shows that only its "local structure" assumption applies to almost all theoretical lengths here (from 0 to about 1024). This also makes it an ideal method to explore the impact of increasing ERLength.\\
        \indent Table~\ref{baseline} suggests our method improves state of the art on Wikia's zero-shot EL dataset. 
        Compared to \citet{logeswaran2019zero}, if we use $E_{repeat}$ to increase the model's ERLength to 1024, we improve the accuracy from 76.06\% to 79.08\%, and for the long data, the improvement is from 74.57\% to 82.14\%. What's more, we also try the Domain Adaptive Pre-training (DAP) method in \citet{logeswaran2019zero}. The combination of DAP and 1024 ERLength raises the result to $80.88\%$.

    \subsection{Impact of increasing ERLlength}
        We further explore the impact of BERT's ERLength on the zero-shot EL task. The red in the table \ref{impactOfERLength} represents the accuracies in the first echelon in each column (for data within a specific DLength interval). It shows a clear step-down trend, which means data with a larger DLength often requires a model with a larger ERLength. What's more, for any column, if we continue to increase the model's ERLength, the accuracy will stabilize within a specific range after the ERLength exceeds most data's DLengths. So the last row in the table is always red, which means that the model with the largest ERLength can always achieve the best level of accuracy on all data of different DLengths.
        
        \begin{figure}[ht]
            \centering
            \includegraphics[width=\linewidth, trim = {0 0 0 0}, clip]{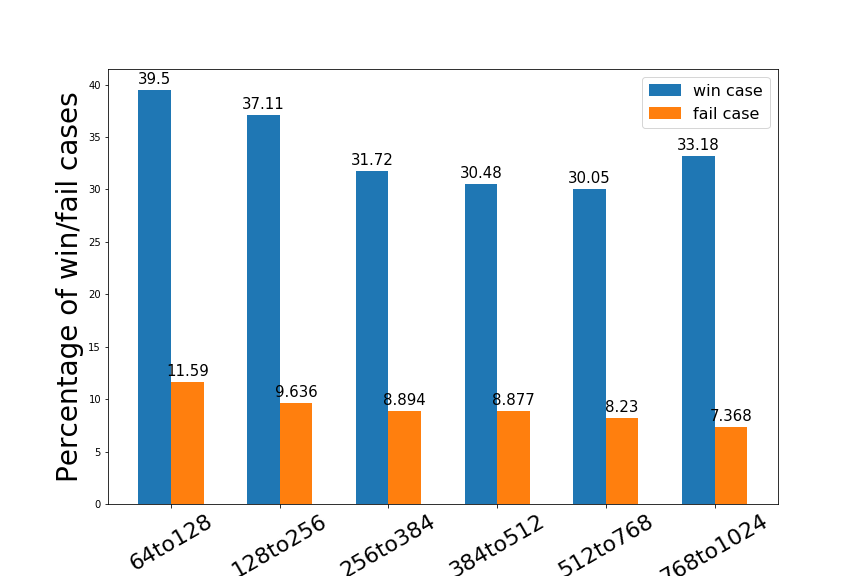}
            \caption{The proportion of win/fail cases during the increase in ERLength. We define \textbf{win} case as the initially wrong data but is now correct after increasing ERLength, and define \textbf{fail} case as the initially correct data is now wrong after increasing ERLength.}
            \label{win&fail case}
        \end{figure}
        
        \indent Figure \ref{win&fail case} shows the changes of win and fail cases when expanding the BERT's ERLength. Generally speaking, when the model can read more content, its accuracy will increase for more valuable information (win case) and decrease for more noise (fail case). The results illustrate that BERT can always use more useful information to help itself while being less disturbed by noise. This once again demonstrates the power of the BERT's full-attention mechanism. This is also the basis on which we can continuously expand BERT's ERLength and continue to benefit. Therefore, for a particular dataset, when we set the ERLength of the BERT, letting it exceed more data's DLength can always bring more improvements.
        
        \begin{figure}[ht]
            \centering
            \includegraphics[width=\linewidth, trim = {0 0 0 0}, clip]{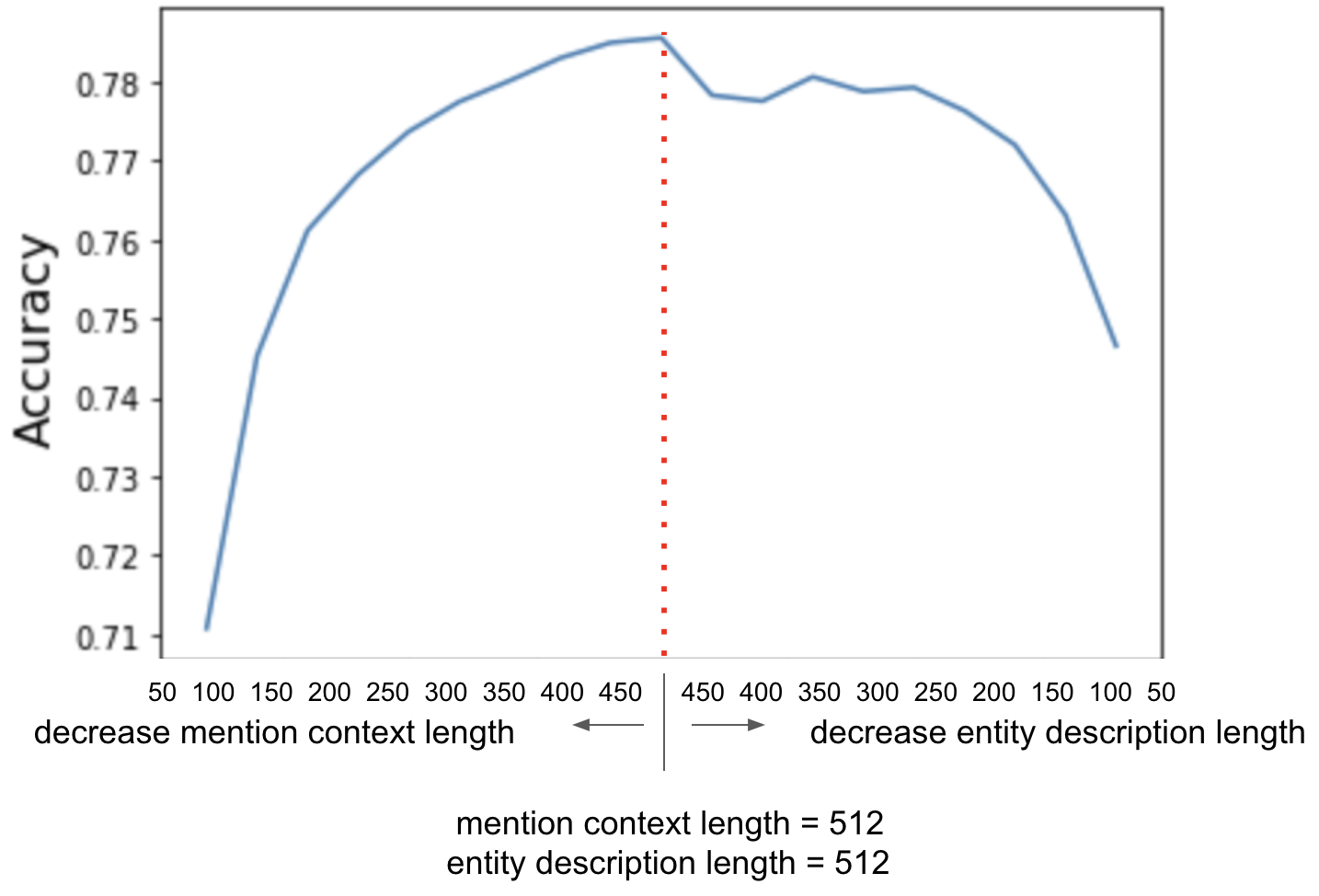}
            \caption{Importance of mention contexts and entity description}
            \label{important_of_mc_ed}
        \end{figure}
        
        \indent Also, in the figure \ref{important_of_mc_ed} we explore the importance of mention contexts and entity descriptions. On Wikia's zero-shot EL dataset, in our settings for BERT with 1024 ERLength, the mention contexts and entity description account for 512, respectively. In figure \ref{important_of_mc_ed}, if we unilaterally reduce the mention contexts and entity description from 512 to 50, the change of accuracy is shown in the figure. It can be found that the two are basically equally important, and no matter which side is reduced, the accuracy will gradually decrease. Therefore, when increasing the BERT ERLength here, the best way is to increase the content of mention contexts and entity description at the same time.

\section{Conclusions and future work}
        We propose an efficient position embeddings initialization method called Embeddings-repeat, which initializes larger position embeddings based on BERT models.  For the zero-shot entity linking task, our method improves the SOTA from 76.06\% to 79.08\% on its dataset. Our experiments suggest the effectiveness of increasing ERLength as large as possible  (e.g., the length of the longest data in the EL experiments). Our future work will be to extend our methods to other NLP tasks.

\section*{Acknowledgments}
        We thank the research supported with Cloud GPUs and TPUs from Google's TensorFlow Research Cloud (TFRC).

\bibliography{emnlp2020.bib}
\bibliographystyle{acl_natbib}

\end{document}